\documentclass{article}

\usepackage[preprint]{neurips_2024}

\usepackage[utf8]{inputenc} 
\usepackage[T1]{fontenc}    
\usepackage{hyperref}       
\usepackage{url}            
\usepackage{booktabs}       
\usepackage{amsfonts}       
\usepackage{nicefrac}       
\usepackage{microtype}      
\usepackage{xcolor}         

\usepackage{graphicx}
\usepackage{amsmath}
\usepackage{amssymb}
\usepackage{mathtools}
\usepackage{amsthm}
\usepackage{multirow}
\usepackage{xspace}
\usepackage{bm}
\usepackage{subfigure}
\usepackage{natbib}
\usepackage{algorithm}
\usepackage{algorithmic}
\usepackage{wrapfig}

\theoremstyle{plain}
\newtheorem{theorem}{Theorem}[section]
\newtheorem{proposition}[theorem]{Proposition}

\theoremstyle{definition}

\theoremstyle{remark}

\setcitestyle{numbers,square}
\newcommand{\ie}{{\emph{i.e.}}\xspace}

\title{Exploiting Fine-Grained Prototype Distribution for Boosting Unsupervised Class Incremental Learning} 

\author{%
  Jiaming Liu, Hongyuan Liu, Zhili Qin, Wei Han, Yulu Fan, Qinli Yang, Junming Shao\thanks{Corresponding Author} \\
    University of Electronic Science and Technology of China\\
    \texttt{\{liujiaming, hongyuanliu, qinzhili, weihan, ylfan\}@std.uestc.edu.cn}\\
    \texttt{\{qinli.yang, junmshao\}@uestc.edu.cn}\\
}

\begin{document}

\maketitle

\begin{abstract}
\label{abstract}

The dynamic nature of open-world scenarios has attracted more attention to class incremental learning (CIL). However, existing CIL methods typically presume the availability of complete ground-truth labels throughout the training process, an assumption rarely met in practical applications. Consequently, this paper explores a more challenging problem of unsupervised class incremental learning (UCIL). The essence of addressing this problem lies in effectively capturing comprehensive feature representations and discovering unknown novel classes. To achieve this, we first model the knowledge of class distribution by exploiting fine-grained prototypes. Subsequently, a granularity alignment technique is introduced to enhance the unsupervised class discovery. Additionally, we proposed a strategy to minimize overlap between novel and existing classes, thereby preserving historical knowledge and mitigating the phenomenon of catastrophic forgetting. Extensive experiments on the five datasets demonstrate that our approach significantly outperforms current state-of-the-art methods, indicating the effectiveness of the proposed method.

\end{abstract}

\section{Introduction}
\label{Sec: Introduction}

Traditional deep learning methods have achieved significant success in many fields \cite{wang2021deep, suganyadevi2022review, wang2022progress, zablocki2022explainability}. However, most of these methods typically operate in static, closed-world scenarios \cite{chen2018lifelong, zhou2022open}, assuming that the whole training data are available at the initial stage without further changing. In contrast, real-world environments are inherently dynamic \cite{parmar2023open}, one of the common characteristics is that categories are evolving and expanding constantly. This reality has led to increased attention in the research of class incremental learning (CIL) \cite{zhou2023deep}, which aims to adapt models to integrate new classes over time.

However, existing CIL methods \cite{rebuffi2017icarl, yu2020semantic, hu2023dense, wang2022learning} often presuppose the availability of all ground-truth labels during training, which is an assumption that does not hold in many real-world scenarios.
With the emergence of data with novel classes, it may be impractical to obtain complete ground-truth labels timely. In a more realistic scenario, only a few labels or even no labels are available. Therefore, this paper focuses on a more realistic and challenging problem of unsupervised class incremental learning (UCIL), aiming to address CIL without any label information.

The most significant challenge of the CIL and UCIL problem is how to capture comprehensive feature representations that effectively model distributions of both existing and incoming unknown classes. This capability is essential for retaining historical knowledge to prevent catastrophic forgetting while also enabling the effective discovery and learning of new classes. Existing CIL methods often rely on strategies including regularization \cite{kirkpatrick2017overcoming, kang2022class}, dynamic structures \cite{li2017learning, serra2018overcoming}, or knowledge rehearsal \cite{jodelet2023class, rolnick2019experience}. These methods utilize supervised data to learn data representation and class decision boundaries with label spaces expanding. However, there is a trade-off for these approaches to incorporate novel concepts and preserve the existing knowledge simultaneously \cite{yan2021dynamically}, especially in an unsupervised manner. In such situations, feature overlap between existing and novel classes adds additional complexity to this problem \cite{wu2022class}. Thus, the primary goal of this paper is to effectively unsupervised modeling knowledge representation. 

Besides, since the ground truth labels are unavailable in UCIL, it brings another challenge to better discover and classify the unknown novel classes effectively. To this end, the key point is to discover potential cluster structures in the feature space and exploit these to learn a classifier. Existing methods often utilize $k$-Means clustering \cite{caron2018deep}, contrastive learning \cite{vaze2022generalized, liu2023open}, or optimal transport-based technique \cite{ntelemis2022information, yang2023bootstrap, liu2023MSciNCD} for unsupervised learning. However, these methods guide the classifier's learning through pseudo-labeling at the category level and may introduce biases and deviations. Therefore, the other goal of this paper is to propose a new strategy for capturing clustering structures in the feature space to boost the discovery and identification of potential unlabeled classes.

In this paper, we introduce a novel framework that exploits fine-grained prototype distribution for boosting UCIL. To capture more comprehensive class representations in an unsupervised manner, we utilize an extensively pre-trained ViT \cite{caron2021dino}. The ViT is used as the feature encoder and remains frozen during training. Besides, we exploit fine-grained prototype distributions to capture more detailed information in the feature space. For more accurate discovery of unlabeled classes, we propose a novel granularity alignment method that guides the training of the classifier through a more fine-grained assignment distribution. Additionally, by capturing the class distribution within prototypes, we develop an overlap reduction strategy between new and existing classes, thereby preserving historical knowledge and mitigating forgetting. Extensive experimental results demonstrate significant improvements over existing state-of-the-art methods, with enhancements of 9.1\% and 9.0\% on the CIFAR100 and CUB datasets, respectively. Our contributions are summarized as follows:

\begin{itemize}
\item We consider a more realistic scenario for class incremental learning in an unsupervised manner and propose an effective approach.
\item We propose a new strategy for modeling detailed class distribution using fine-grained prototypes and boosting more accurate class discovery through granular alignment.
\item We achieve significant performance gains over state-of-the-art methods on five benchmark datasets. Extensive case studies demonstrate the effectiveness of the proposed method from various perspectives.
\end{itemize}

\section{Related Works}

\textbf{Class Incremental Learning.} 
In the realm of class incremental learning, the primary challenge is to learn new classes while retaining knowledge of previously learned ones, across tasks with non-overlapping class sets. In addressing this challenge, three primary strategies have emerged including rehearsal-based methods \cite{rebuffi2017icarl, rolnick2019experience, liu2020mnemonics, jodelet2023class}, regularization-based methods \cite{kirkpatrick2017overcoming, titsias2019functional, yu2020semantic, kang2022class}, and dynamic architecture methods \cite{li2017learning, serra2018overcoming, yan2021dynamically, hu2023dense}.
A notable trend in recent CIL research involves the integration of pre-trained transformer networks as feature extractors \cite{wang2022learning, mcdonnell2023ranpac, zhang2023slca}, enhancing better representation and reducing forgetting. However, these existing methods require label information at each incremental step, which is not feasible in real-world scenarios. Consequently, a new challenge lies in how to effectively discover categories within continuous streams of unlabeled data. One recent solution is Msc-iNCD \cite{liu2023MSciNCD}, which utilizes cosine normalization of classifier weights alongside a frozen feature mechanism. Building upon this foundation, our method introduces a novel method, that offers a new perspective in addressing the challenges of CIL without labels.

\textbf{Novel Class Discovery.} 
The task of novel class discovery (NCD) focuses on identifying unknown classes within an unlabeled dataset with the help of the known labeled dataset. Traditional approaches to NCD \cite{han2020automatically, fini2021unified, han2019learning} typically operate under the unrealistic assumption that the unlabeled set only samples from unknown novel categories. To improve, generalized class discovery (GCD) \cite{vaze2022generalized, pu2023dynamic, liu2023open} tackles a more complex scenario where a test set consists of both known and unknown classes, all of which need to be identified concurrently. Several recent advancements have focused on incrementally discovering and learning classes in a sequential manner \cite{wu2023metagcd, roy2022class, joseph2022novel, liu2022residual}. These methods, however, are constrained by several factors, which limit the number of incremental learning sessions \cite{joseph2022novel}, require an identifier for each task (task-id) \cite{liu2022residual}, and necessitate access to labeled data in every session \cite{wu2023metagcd}. Such constraints limit their applicability to dynamic, real-world environments where unknown classes evolve continuously. Our proposed method is designed to offer a more adaptable and realistic approach to NCD in the unlabeled continuous class sequence.

\section{Proposed Method}
\label{sec: method}

\subsection{Problem Definition}

\begin{figure}[tp]
\centering \hspace{0mm}
\includegraphics[width = 1\textwidth]{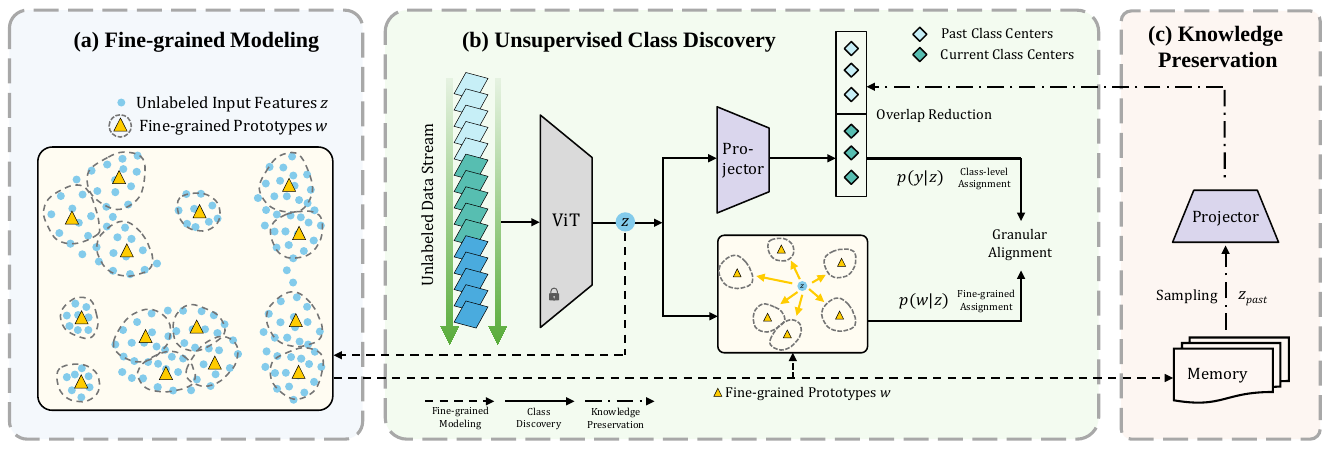}
\caption{An illustration of the proposed method, including three parts: (a) Fine-grained modeling, which explores fine-grained prototype distributions to capture more detailed information in the feature space. (b) Unsupervised class discovery, which guides the training of the classifier by granularity alignment. (c) Knowledge preservation, which extracts historical class distribution and decreases overlaps between new and existing classes.}
\label{fig:framework}
\end{figure}

Here we introduce the problem setup of unsupervised category incremental learning (UCIL). In this setting, the model continuously observes $\mathcal{T}$ sequential tasks, each of which presents the data set $\mathcal{D}_{t}$ with only unlabeled samples $x^t_i \in \mathcal{X}_{t}$. The corresponding label sets $\mathcal{Y}_t$ can not be assessed during training. 
As the common settings in CIL field \cite{rebuffi2017icarl,liu2023MSciNCD}, classes from different tasks are disjoint, \ie, $\mathcal{Y}^{[\mathrm{i}]} \cap \mathcal{Y}^{[j]}=\emptyset$, and model can only acquire the classes of the current task. We assume the number of novel classes $cNum = |\mathcal{Y}_t|$ is pre-known. At each task $t$, our task aims to discover unknown categories in $\mathcal{D}_{t}$ and identify the instances well into the discovered clusters, as well as maintain the performance on the previous classes from $\mathcal{D}_{1}$ to $\mathcal{D}_{t-1}$. 
In other words, the proposed method incrementally and learns the classifier that can group any test instance $x$ into the categories $\mathcal{X} \rightarrow \bigcup_{t=1}^\mathcal{T} \mathcal{Y}_{t}$, which are discovered from the unlabelled data steams $\mathcal{D}=\left\{\mathcal{D}_{1}, \mathcal{D}_{2}, \cdots, \mathcal{D}_{\mathcal{T}}\right\}$, without the help of task-id.

\subsection{Framework Overview}

Before introducing the detailed method, we first present the network architecture of our method, which is illustrated in Fig. \ref{fig:framework}. Our method adopts a pre-trained vision transformer ViT-B/16 \cite{caron2021dino} as the feature extractor $f$ for more semantic features $z$. The parameters of $f$ are frozen in all training stages to avoid forgetting and reduce the amount of parameter calculations. Following the feature extractor, trainable prototypes $w$ are used for modeling the fine-grained distribution in the feature space. The other branch contains a classifier composed of a projector $g(\cdot)$ and class centers $c$ for class identification and prediction. The knowledge of historical classes can be stored using these prototypes and then sampled to reduce the overlap with new classes.

\subsection{Distribution Modeling by Fine-grained Prototypes}

The primary objective for UCIL is to capture comprehensive feature representations. Previous methods \cite{liu2023MSciNCD, roy2022class} typically utilize a single prototype per category and store exemplars for knowledge preserving, based on the assumption of single Gaussian-like class distributions. In contrast, this paper addresses this challenge by exploiting fine-grained prototypes to capture detailed distributions in the feature space. By effectively modeling the distributions of both existing and coming classes, this approach not only mitigates knowledge forgetting but also enhances the discovery and learning of new classes.

\begin{wrapfigure}[22]{r}{0.33\textwidth}
\begin{minipage}{0.33\textwidth}
\vspace{-24pt}
\begin{figure}[H]
    \centering
    \subfigure{
        \includegraphics[width=0.98\textwidth]{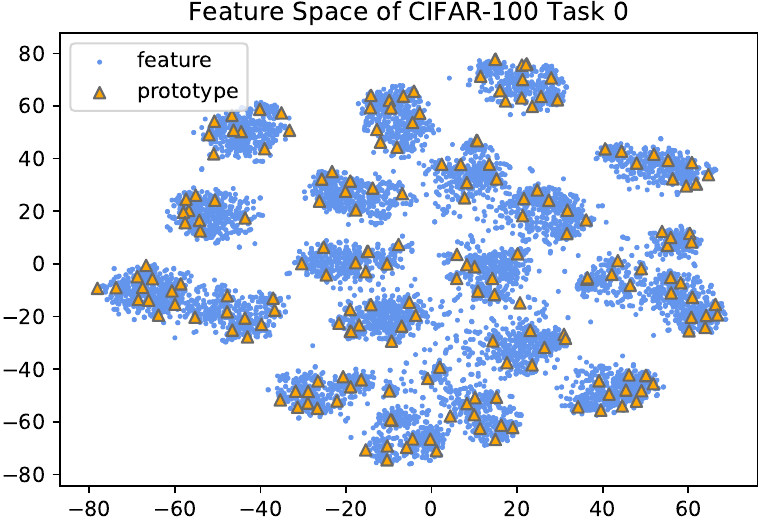}
    }
    \hspace{-10mm}
    \subfigure{
	\includegraphics[width=0.98\textwidth]{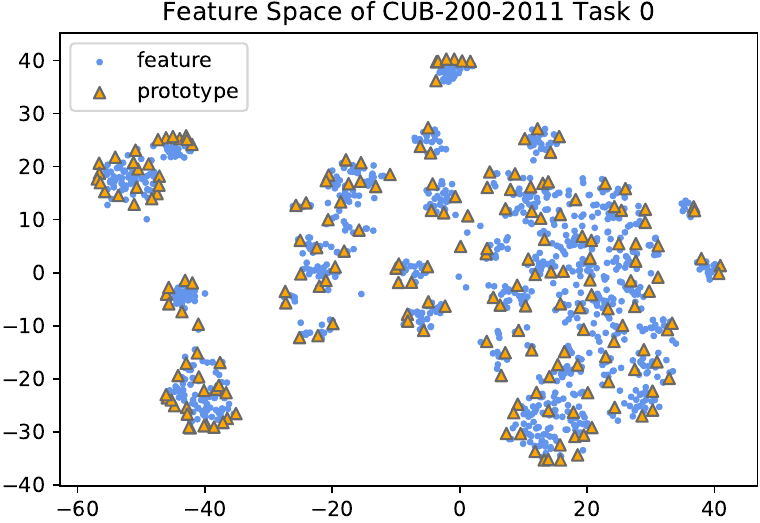}
    }
    \vspace{-4mm}
\caption{The visualization of fine-grained Gaussian prototype distribution within feature space.}
\label{fig:tsne}
\end{figure}
\end{minipage}
\end{wrapfigure}

To offer a more illustrated understanding, we provide a visualization of these prototypes within the feature space in Figure \ref{fig:tsne}. In this figure, the triangle represents the fine-grained prototype, there are 20 and 40 real classes in the up and down figures, respectively. We can see that the learned prototypes can capture comprehensive feature representations, and mimic the density of feature distribution, capturing the sub-cluster structures of the feature distribution.

In greater detail, the fine-grained prototypes can be explored and optimized based on the in a Gaussian mixture manner. Formally, for an input data $x_i\in \mathcal{D}_t$ in task $t$, we can obtain the feature by $z_i = f(x_i)$. 
Besides, our method adopts a Gaussian prototypes set $\mathcal{W}$ in the feature space, where $w \sim  \mathcal{N}(\mu_w, \sigma^{2}_{w}I)$. Here, $\mu_w$ and $\sigma_w$ are trainable parameters. To model the undefined distribution in the feature space at a fine-grained level, we associate each feature $z_i$ to these fine-grained Gaussian prototypes to indicate to which fine-grained subclass that $z_i$ belongs. Suppose $r=pNum_{t}$ is the number of prototypes in task $t$, which is pre-defined and far more than the potential class number. Assume that $w_i$ and $z_i$ are $l_2$-normalized, then posterior probability $p(w_i|z_i)$ is given by:
\begin{equation}
\label{eq: p(w|z)}
     p_{}\left(w_{i} \mid {z}_{i}\right)=
 \frac{\exp \left[2({z}_{i}^{\top} \cdot {\mu}_{w_{i}} -1 )/ \sigma_{w_{i}}^{2}\right] \pi_{w_{i}}}{\sum_{w_{i} \in \mathcal{W}  } \exp \left[2({z}_{i}^{\top} \cdot {\mu}_{w_{i}^{\prime}} -1)/ \sigma_{w_{i}^{\prime}}^{2}\right] \pi_{w_{i}^{\prime}}},
\end{equation}
where $\pi_{w_{i}}$ is the weight of Gaussian prototype and $\sigma_{w_{i}}$ is in diagonal form for simplification. However, it is hard to directly optimize by minimizing the negative log-likelihood on $n$ observed features, and hence we resort to a variational approximation $q(w|z)$ to true distribution $p(w|z)$. The evidence lower bound (ELBO) of this problem is as follows:

\begin{equation}
\label{eq: vi}
\begin{aligned}
\text{ELBO}
& =\frac{1}{n} \sum_{i=1}^{n} \log \left[\sum_{w_{i}=1}^{r}  p\left(w_{i} \mid {z}_{i}\right) \right]  
 =\frac{1}{n} \sum_{i=1}^{n} \log \left[\sum_{w_{i}=1}^{r} q\left(w_{i} \mid {z}_{{i}}\right) \frac{p\left(w_{i} \mid {z}_{i}\right)}{q\left(w_{i} \mid {z}_{{i}}\right)}\right] && \\
& \geq \frac{1}{n} \sum_{i=1}^{n} \sum_{w_{i}=1}^{r} \left[\right. q\left(w_{i} \mid {z}_{{i}}\right) \log p \left( w_{i} \mid {z}_{i}\right)   - q\left(w_{i} \mid {z}_{i}\right)\log q\left(w_{i} \mid {z}_{{i}}\right)\left.\right], \\
\end{aligned}
\end{equation}
which can be maximized by alternately inferring the posterior $q(w|z)$. Inspired by \cite{ni2021superclass}, the problem can be solved through an Expectation-Maximization (EM) algorithm.

\textbf{E-step. }
Given a certain $p(w|z)$, to infer $q(w|z)$, let $\mathbf{Q}_{w_{i}, i}=q\left(w_{i} \mid {z}_{i}\right)$ and $\mathbf{P}_{w_{i}, i}=p\left(w_{i} \mid {z}_{i}\right)$ be two $r \times n$ matrices. To increase diversity and avoid degradation, we enforce that each fine-grained Gaussian component is assigned an equal number of samples, where $\pi_{w_{i}}$ is equal for all $i$. Then we can covert the E-step to an optimal transport problem as
\begin{equation}
\label{eq: optimal transport}
         \min _{\mathbf{Q} \in \mathcal{Q}}-\left(\operatorname{Tr}\left(\mathbf{Q}^{\top} \log \mathbf{P}\right)+ \epsilon H(\mathbf{Q})\right) 
\end{equation}
\begin{equation}
 \mathcal{Q}=\left\{\mathbf{Q} \in \mathbb{R}_{+}^{r \times n} \left\lvert\, \mathbf{Q} \mathbf{1}_{n}=\frac{1}{r} \mathbf{1}_{r}\right., \mathbf{Q}^{\top} \mathbf{1}_{r}=\frac{1}{n} \mathbf{1}_{n}\right\},
\end{equation}
where $H(\mathbf{Q})=-\sum_{i j} \mathbf{Q}_{i j} \log \mathbf{Q}_{i j}$ is the entropy function, $\mathbf{1}_{n}$ is the vector of ones in dimension $n$. $\epsilon$ is the scaling parameter to control the smoothness of the assignment.

The optimal solution of Eq. \ref{eq: optimal transport} can be written as $\mathbf{Q}^{*}=\operatorname{Diag}(\mathbf{u}) \mathbf{P}^{\lambda} \operatorname{Diag}(\mathbf{v})$, where $\mathbf{u} \in \mathbb{R}^{r}$ and $\mathbf{v} \in \mathbb{R}^{n}$ are renormalization vectors. 
In practice, the problem can be efficiently solved by a small number (we set it to 3) of matrix multiplications employing the Sinkhorn-Knopp \cite{cuturi2013sinkhorn} algorithm.

\textbf{M-step. } 
After obtaining the optimal $q^{\ast}(w|z)$ in E-step, we can optimize trainable $\mu$ and $\sigma$, by gradient methods as:
\begin{equation}
\label{eq: m-step}
\mathcal{L}_{proto}=-\frac{1}{n} \sum_{i=1}^{n} q^{\ast}\left(w_{i} \mid {z}_{i}\right)\log p\left(w_{i} \mid {z}_{i}\right) 
\end{equation}

\textbf{Discussion: }
Eq. \ref{eq: m-step} shares a similar formulation with prototypical contrastive learning methods, such as SwAV \cite{caron2020swav}. However, while these methods primarily focus on learning representations, our approach places theoretical emphasis on these fine-grained prototypes, which are formed as Gaussians rather than a single vector. A case study is conducted on it in the experiment part. Fine-grained prototype modeling offers two key benefits: firstly, it helps retain task-specific knowledge for overlap reduction between existing and new classes, thereby preventing forgetting. Secondly, this method enables the capturing of more refined feature distributions, which will enhance the unsupervised category discovery and will be discussed in the following section.

\subsection{Unsupervised Class Discovery by Granular Alignment} 

In this section, we will introduce a novel granular alignment strategy to discover the unknown class by the explored fine-grained Gaussian prototypes. To begin with, given the extracted features $z=f(x)$, the classifier is defined as:
\begin{equation}
Y = p(y|z) =  \frac{\text{exp} \left \langle   c^y \circ g(z) / \tau \right \rangle }{\sum_{c' \in \mathcal{C}^t  } \text{exp} \left \langle c'\circ g(z) / \tau \right \rangle},
\label{eq: p(y|z)}
\end{equation}
where 
$\mathcal{C}^t = \left \{c^{t}_1 ... c^{t}_{k}\right \}$ is the trainable class center for each discovered category and $k$ is the class number in current task. $\tau$ is the temperature parameter. $g(z)$ is the projector to transform $z$ into a reduced space which focuses on the course-level cluster discovery task. Following previous methods, $\left \langle \circ \right \rangle $ is the cosine distance by $l_2$ normalized on each $c$ and $g(z)$.

The previous section has obtained the fine-grained feature assignment probability on multiple Gaussians $W=p(w|z)$. Then, learn the classifier in an unsupervised manner, the previously optimized fine-grained assignment probability $\mathbf{Q}$ can be regarded as the guidance to learn the classifier. In the implementation, we align the granularity by maximizing the mutual information between class-level assignment probability $Y$ and fine-grained prototype assignment probability $W$ as:
\begin{equation}
\begin{aligned}
 \mathcal{L}_{align}= -I(W ; Y) 
=\sum_{(w, y)} p(w, y) \log \left(\frac{p(w, y)}{p(w)}\right)
-\lambda_{GA}\sum_{y} p(y) \log (p(y)),
\end{aligned}
\end{equation}
where $\lambda_{GA}$ is the regularization weight, $p(w, y)$ is the joint probability, and $p(w)$ and $p(y)$ are marginal terms.

\begin{proposition} \textit{$p(w, y)$ is equal to the matrix multiply by $W^{T} \cdot Y  \in \mathbb{R}^{r\times k} $}. 
\end{proposition}

\textit{Proof. } 
$W=p(w|z)\in \mathbb{R}^{n\times r}$ and $Y=p(y|z)\in \mathbb{R}^{n\times k}$ are conditional independent on $z$ as they are computed by independent network parameters, then
\begin{equation}
\begin{aligned}
p\left(w_{j}, y_{i}\right) 
 \approx \sum_{z}p\left(w_{j},y_{i} \mid z\right)
 = \sum_{z} p\left(w_{j} \mid z\right) \cdot p\left(y_{i} \mid z\right)
 =\left(W^{T} \cdot Y\right)_{i j}. 
\end{aligned}
\end{equation}
Based on the proof, we can also approximate the conditional assignment relationship $p(y|w)$ from each Gaussian prototypes to the class centers by:
\begin{equation}
\label{eq: p(y|w)}
p(y|w)=\frac{p(w,y)}{p(w)} =\frac{p(w,y)}{\sum_{y}p(w,y)}, 
\end{equation}
which can be utilized for knowledge preservation for the detailed class distribution during incremental learning, with further details provided in the subsequent section.

\textbf{Discussion:} 
Most existing methods \cite{liu2023MSciNCD, rizve2022towards} primarily guide the unsupervised learning of the class discovery or clustering predictors by utilizing pseudo labels at the same coarse class level. By employing granular alignment between $Y$ with $W$, the clustering model is encouraged to learn richer feature representations and incorporate these captured more details with fine-grained assignment distribution $W$. This approach leads to more informed clustering decisions thereby enhancing the unsupervised class discovery performance at the coarser class level $Y$. The benefits of this method will be demonstrated in the experimental section.

\subsection{Knowledge Preservation by Overlap Reduction}
In the previous section, we focus on unsupervised class discovery within a specific task $t$. Without history knowledge, this approach might risk feature overlapping between old class centers $\mathcal{C}_{old}=\left \{ \mathcal{C}_{1} \cdots \mathcal{C}_{t-1} \right \}$ and current ones $\mathcal{C}_t$, potentially leading to forgetting. To mitigate this issue, this section introduces an overlap reduction technique for fine-tuning the classifier, which enables the identification of both current and historical classes without task-id information without forgetting.

At the end of each task, the knowledge of the current task can be modeled and stored by the fine-grained Gaussian prototypes set $\mathcal{W}$. In detail, firstly, obtain the corresponding class label of each Gaussian prototype by $p(y|w)$ as Eq. \ref{eq: p(y|w)}. Then, derive the assignment probability of each feature by $p(w|z)$ as Eq. \ref{eq: p(w|z)}. Therefore, we can compute the statistic information of count $n_{w}$, the mean $\mu_w$ and variance ${\sigma}_{w}^{2}\mathit{I}$ by the sample features assigned to each $w$. We can also compute the purity $\textsc{pur}_{w}$ by the ratio of the majority class sample size and total sample size assigned to each Gaussian. The recalculates the parameters of each Gaussian can better fit the current feature distribution.

Therefore, for a new task, the knowledge of old classes can be derived by sampling features from the memorized prototype set:
\begin{equation}
\mathcal{D}_{old} = \left \{ z,y \right \} \sim \sum_{w\in\mathcal{W}^y } \pi_{w} \mathcal{N}\left(z \mid {\mu}_{w}, {\sigma}_{w}^{2}\mathit{I} \right),
\label{eq:sampling}
\end{equation}
where $\pi_{w} = n_{w} \times \textsc{pur}_{w} / \sum_{w'\in\mathcal{W}^y }n_{w'} \times \textsc{pur}_{w'}$ is the weight term for each $w$. We sample equal sizes of features for each class $y$ to ensure class balance.

To avoid forgetting, we aim to maintain the ability to discriminate the old classes in the current classifier with joint class centers $c \in \mathcal{C}_{old} \cup \mathcal{C}_t $ by a cross-entropy-like objective function:
\begin{equation}
    \mathcal{L}_{old}=\frac{1}{\left|\mathcal{D}_{old} \right|} \sum_{\genfrac{}{}{0pt}{}{(z, \hat{y}) \in \mathcal{D}_{old}}{ c^{y} \in \mathcal{C}_{ old}} }\frac{\exp \left \langle   c^y \circ g(z) / \tau \right \rangle }{\sum_{c' \in \mathcal{C}_{old} \cup \mathcal{C}_t } \exp \left \langle c' \circ g(z) / \tau \right \rangle}.
\label{eq:oldloss}
\end{equation}

Besides, to solve the feature overlapping between current classes and old classes, we proposed a separation loss to make the trainable class centers and features in the current task far away from those of old tasks. This is achieved by encouraging features from the current task $z \in \mathcal{D}_{ t}$ to generate higher output logits on current class centers $\mathcal{C}_{t}$ rather than centers of old classes $\mathcal{C}_{old}$:
\begin{equation}
\mathcal{L}_{sep}= \frac{1}{\left|\mathcal{D}_{t} \right|} \sum_{z \in \mathcal{D}_{t}}\frac{\sum_{c \in \mathcal{C}_{t}}\text{exp} \left \langle   c \circ g(z) / \tau \right \rangle }{\sum_{c' \in \mathcal{C}_{old} \cup \mathcal{C}_t } \text{exp} \left \langle c' \circ g(z) / \tau \right \rangle}.
\label{eq:seploss}
\end{equation}

Then the final overlap reduction object is formed as:
\begin{equation}
\mathcal{L}_{reduct} =  \lambda_{old}\mathcal{L}_{old} + \mathcal{L}_{sep}.
\label{eq:jointclassifier}
\end{equation}

\subsection{Training Procedure}

\begin{wrapfigure}[14]{rT}{0.52\textwidth}
    \begin{minipage}{0.52\textwidth}
    \vspace{-26pt}
        \begin{algorithm}[H]
        \renewcommand{\algorithmicrequire}{\textbf{Input:}}
        \renewcommand{\algorithmicensure}{\textbf{Output:}}
          \caption{Training Procedure for Each Task.}\label{alg:algorithm1}
            \small
            \begin{algorithmic}[1]
            \REQUIRE $\{\mathcal{D}_t\}_{t=1}^{\mathcal{T}}$, number of Gaussian prototypes $pNum$, number of class $cNum$, memory $\mathcal{M}$ \\
            \FOR{each mini-batch in $\mathcal{D}_t$}
                \STATE Obtain $\mathbf{Q}$ by Eq. \ref{eq: optimal transport} 
                \STATE Compute $\mathcal{L}_{proto}$ and $\mathcal{L}_{align}$ with the obtained $\mathbf{Q}$ 
                \STATE Sampling $\mathcal{D}_{old}$ from memory $\mathcal{M}$ by Eq. \ref{eq:sampling}
                \STATE Compute $\mathcal{L}_{reduct}$ by Eq. \ref{eq:all}
                \STATE $\mathcal{L} \leftarrow \mathcal{L}_{proto} + \mathcal{L}_{align} + \mathcal{L}_{reduct}$
                \STATE Update $\mathcal{W}_t, \mathcal{C}_t, g_t$ by $\text{AdamOptimizer}(\mathcal{L})$
            \ENDFOR
            \STATE Retrieval $\mathcal{M}_t$ by updated $\mathcal{W}, \mathcal{C}, g$
            \STATE $\mathcal{M} \leftarrow \mathcal{M} \cup \mathcal{M}_t$ 
          \end{algorithmic}
        \end{algorithm}
    \end{minipage}
\end{wrapfigure}

The overall training objective function is:
\begin{equation}
\mathcal{L}  = \mathcal{L}_{proto} + \mathcal{L}_{align} + \mathcal{L}_{reduct}.
\label{eq:all}
\end{equation}
Specifically, as shown in Fig. \ref{fig:framework} and Alg. \ref{alg:algorithm1}, the training learning process can be decoupled into the following parts.

\textit{(1) Fine-grained Prototype Modeling.}
Capture the detailed feature distribution using fine-grained Gaussian prototype modeling, where the number of Gaussian prototypes $pNum$ significantly exceeds the number of categories $cNum$. 
\textit{(2) Unsupervised Class Discovery.}
Guide the training of the classifier by maximizing the granular alignment between fine-grained prototype assignment $p(w|z)$ and class-level assignment $p(y|z)$.
\textit{(3) Preserving History Knowledge.}
Store the rich statistic information for features $z$ on the learned Gaussian prototypes $w$. Sample old data $\mathcal{D}_{old}$ from stored fine-grained prototype and concatenate with the current data $\mathcal{D}_{t}$ to fine-tune the classifier, thereby avoiding feature overlap and preventing forgetting.

\section{Experiments}

\subsection{Experimental Settings}
\label{sec: settings}

\textbf{Datasets and Splits.} 
To demonstrate the effectiveness of our method, we conducted experiments on five benchmark datasets: CIFAR-10 (C10) \cite{krizhevsky2009learning}, CIFAR-100 (C100) \cite{krizhevsky2009learning}, Tiny ImageNet (T200) \cite{le2015tiny}, CUB200-2011 \cite{WahCUB_200_2011}, and Herbarium-683 (H683) \cite{tan2019herbarium}. Notably, each benchmark training dataset is divided into either two or five subsets, corresponding to the two-step and five-step settings, with non-overlapping label spaces. For example, in the five-step setting, CIFAR-100 is partitioned into 5 tasks, with each task encompassing 20 classes. Evaluation is performed on the test dataset.

\textbf{Evaluation Protocol.}  
We follow the evaluation protocol outlined in Msc-iNCD \cite{liu2023MSciNCD}. Our primary reporting metric is the clustering accuracy after the final continual learning session, referred to as the overall discovery accuracy $\mathcal{A}$. This metric is assessed on a test set comprising instances from all previously encountered classes, and the assignment is determined using the Hungarian algorithm \cite{kuhn1955hungarian}. Furthermore, we also measure the degree of forgetting by the forgetting score $\mathcal{F}$, which quantifies the drop in clustering accuracy between the first session $\mathcal{T}_1$ and the last session $\mathcal{T}{\text{end}}$ especially for the classes $\mathcal{Y}_1$ discovered in the initial session $\mathcal{T}_1$. It is important to note that, during inference, the task ID remains unknown in all methods. In certain specific experiments, we also employ task-specific discovery accuracy to assess the performance of discovering new classes within a single session.

\textbf{Implementation Details.}
We employ a ViT-B/16 \cite{dosovitskiy2020vit} as the backbone $g$, utilizing pre-trained weights from DINO \cite{caron2021dino}, which are kept frozen throughout all experiments. The 768-dimensional output [CLS] token serves as the feature vector denoted as $z$. The number of fine-grained prototypes is set to $pNum=1000$. For projection, we utilize a 2-layer MLP with 768 hidden units and an output dimension of 128. We use the default hyperparameters for the Sinkhorn-Knopp algorithm \cite{cuturi2013sinkhorn}, setting $\epsilon$ to 0.05 and $n_{iter}$ to 3. Each session undergoes 200 epochs of training, with a batch size set at 512. Adam is employed as the optimizer with a default learning rate of 1e-3. Additionally, we assign a weight of 10 to the old loss $\lambda_{old}$ and a weight of 4 to the GA regularization term $\lambda_{GA}$. The default temperature parameter $\tau$ is maintained at 0.1. We will later explore the sensitivity of these hyperparameters in our experiments.
The experiments were conducted on a server with an Intel(R) Xeon(R) E5-2678 2.50GHz CPU, an NVIDIA GeForce RTX 3090 GPU, and 32GB of memory.

\subsection{Comparison with Leading Approaches}

\textbf{Compared Baselines.}
We have selected seven baseline methods for comparative analysis:
The first five are extended baselines. Two of them are related incremental novel class discovery (iNCD) methods including \textit{ResTune} \cite{liu2022residual} and \textit{FRoST} \cite{roy2022class}, where \textit{ResTune} discovers the novel classes without labels but with task-id and \textit{FRoST} needs supervised information to discover classes incrementally. In addition, several representative CIL methods are extended for comparison including two regularization-based methods \textit{EwC} \cite{kirkpatrick2017overcoming}, \textit{LwF} \cite{li2017learning}, and a rehearsal-based method \textit{DER} \cite{yan2021dynamically}. 
While the discussed methods were originally developed under fully supervised conditions, we have adapted them to fit the UCIL framework in our approach. In this adaptation, real labels are substituted with pseudo-labels, obtained through the same self-training strategy as employed in \textit{Msc-iNCD}. All adapted methods use DINO pre-trained ViT-B/16 \cite{caron2021dino} as the backbone. In our evaluations, we adhere to the same protocol and primarily compare our method with \textit{Msc-iNCD} \cite{liu2023MSciNCD} and its advanced version \textit{Msc-iNCD++} \cite{liu2023MSciNCD}. Although there is no publicly available code for \textit{MSc-iNCD}, we simply reproduce their method by following the pseudo-code provided in their paper, which achieves the reported performance levels. We report the results of the rest of the baseline methods as outlined in the \textit{MSc-iNCD} paper, as we have maintained the same experimental setup.

\textbf{Results.}
Tab. \ref{tab:baselines} showcases the comparison results with adapted baseline methods. As can be observed, extended baselines do not perform well in the UCIL task. For instance, \textit{DER} exhibits instability due to cumulative errors learning from incorrect pseudo-labels, while \textit{ResTune} struggles in this task as it relies on task-id information. These results underscore the challenges associated with the UCIL task. Furthermore, our method consistently outperforms the protocol methods \textit{MSc-iNCD} and \textit{MSc-iNCD++} by a substantial margin across various datasets. In the more demanding five-step setting, the performance gap between our method and others widens progressively, indicating that our approach excels in challenging scenarios with longer steps. Notably, in the incremental five-step setting, we achieve significant improvements in overall discovery accuracy, enhancing it by 9.1\% and 9\% on the CIFAR-100 and CUB datasets, respectively.

\begin{table*}[!t]
\setlength{\tabcolsep}{5pt}
\caption{Comparison with baseline methods in five datasets on two types of incremental splits. The overall discovery accuracy is reported.  The best method in each dataset is emphasized in \textbf{bold}. 
}
\vspace{-3mm}
\label{tab:expt_sota}
\begin{center}
\resizebox{\textwidth}{!}{
\begin{tabular}{lcccccc|cccccc}
\toprule               
\multirow{2}{*}{Methods}& \multicolumn{6}{c}{Two-step} & \multicolumn{6}{c}{Five-step}\\
        & C10 & C100 & T200 & B200 & H683 & AVG & C10 & C100 & T200 & B200 & H683 & AVG  \\
\midrule  
EwC \cite{kirkpatrick2017overcoming}       & 79.0 & 43.9 & 33.3 & 25.5 & 25.1 & 41.4 & 81.1 & 30.6 & 23.2 & 19.1 & 22.4 & 35.3 \\
LwF \cite{li2017learning}       & 34.4 & 42.4 & 27.2 & 23.9 & 24.9 & 30.6 & 25.8 & 16.1 & 15.6 & 15.7 & 23.4 & 19.3 \\
DER \cite{yan2021dynamically}       & 69.9 & 30.3 & 28.9 & 20.4 & 24.7 & 34.9 & 76.2 & 36.2 & 21.7 & 16.3 & 22.3 & 34.5 \\
ResTune \cite{liu2022residual}   & 47.2 & 17.1 & 17.2 & 13.0 & 17.1 & 22.3 & 49.2 & 19.4 & 12.2 & 12.4 & 11.2 & 20.9 \\
FRoST \cite{roy2022class}     & 46.6 & 34.2 & 26.1 & 17.6 & 18.4 & 28.6 & 69.2 & 43.6 & 31.0 & 18.5 & 23.4 & 37.1 \\
MSc-iNCD \cite{liu2023MSciNCD}  & 89.2 & 60.3 & 54.6 & 28.7 & 25.7 & 51.7 & 85.4 & 63.7 & 53.3 & 28.9 & 25.2 & 51.3 \\
MSc-iNCD++ \cite{liu2023MSciNCD} & 90.9 & 61.4 & 55.1 & 36.9 & 27.5 & 54.4 & 91.7 & 67.7 & 56.5 & 41.1 & 26.1 & 56.7 \\ 
\midrule 
\multirow{2}{*}{Ours} & \textbf{91.8} & \textbf{68.9} & \textbf{57.8} & \textbf{45.0} & \textbf{27.7} & \textbf{58.2} & \textbf{94.7} & \textbf{76.8} & \textbf{63.3} & \textbf{50.1} & \textbf{27.5} & \textbf{62.5} \\
          & (+0.9) & (+7.5) & (+2.7) & (+8.1) & (+0.2) & (+3.8) & (+3.0) & (+9.1) & (+6.8) & (+9.0) & (+1.4) & (+5.8) \\
\bottomrule
\end{tabular}
}
\end{center}
\vspace{-8mm}
\label{tab:baselines}
\end{table*}

\subsection{Case Study for Each Component}

In this section, we perform a series of case studies across different scenarios and perspectives, aiming to assess the individual contributions of each component within our method. 

\begin{wraptable}[12]{rT}{0.45\textwidth}
\begin{minipage}{0.45\textwidth}
\begin{table}[H] 
\centering
\scriptsize
\setlength{\tabcolsep}{5pt}
\vspace{-22pt}
\caption{Performance on unsupervised class discovery in each specific static task.}
\vspace{-1.3mm}
\label{tab:MI}
\resizebox{\textwidth}{!}{
\begin{tabular}{lccc}
\toprule
Methods & C100 & T200 & B200 \\
\midrule
$k$-Means      & 77.8   $\pm$ 4.7 & 70.4   $\pm$ 1.6 & 53.9  $\pm$ 4.8 \\
MSc-iNCD     & 81.5   $\pm$ 6.6 & 75.0   $\pm$ 2.7 & 54.3  $\pm$ 5.6 \\
MSC-iNCD+GA  & 86.4   $\pm$ 4.8 & 77.9   $\pm$ 2.5 & 54.7  $\pm$ 2.7 \\
\midrule
Ours (w/o GA)& 85.5   $\pm$ 5.8 & 77.0   $\pm$ 4.9 & 54.9  $\pm$ 3.7 \\
Ours (p500)  & 86.9   $\pm$ 5.5 & 77.4   $\pm$ 3.5 & 59.1  $\pm$ 2.9 \\
Ours (p2000) & 89.8   $\pm$ 4.1 & 79.3   $\pm$ 1.1 & 60.3  $\pm$ 3.0 \\
Ours (p5000) & \textbf{91.0   $\pm$ 3.2} & \textbf{81.8   $\pm$ 2.5} & \textbf{60.5  $\pm$ 3.1} \\
\midrule
Ours (w/o $\sigma$) & 89.2   $\pm$ 5.8 & 76.9   $\pm$ 2.6 & 56.6  $\pm$ 3.0 \\
Ours (p1000) & 90.2   $\pm$ 3.0 & 79.1  $\pm$ 2.1 & 59.3  $\pm$ 3.2 \\
\bottomrule
\end{tabular}
}
\end{table}
\end{minipage}
\end{wraptable}

\textbf{Granular Alignment for Unsupervised Class Discovery.}
In this section, we highlight the effectiveness of the granular alignment strategy as self-guidance in unsupervised class discovery tasks. Our focus here is solely on the performance of task-specific class discovery within each session, irrelevant to incremental scenarios. To quantify this, we measure both the mean and variance of clustering accuracy across 5 individual tasks, and the results are presented in Tab. \ref{tab:MI}. We compare our proposed method with \textit{$k$-Means}, \textit{MSc-iNCD}, and an adapted version of \textit{MSc-iNCD} equipped with our proposed Granular Alignment (GA) method (referred to as \textit{MSc-iNCD+GA}). Additionally, we explore the impact of removing GA from our proposed method, achieved by setting one prototype per class (referred to as \textit{ours w/o GA}). Furthermore, we examine our method's performance with varying numbers of fine-grained prototypes. In addition, we also experiment with fixing the trainable parameter $\sigma$ in Eq. \ref{eq: p(w|z)}, naming it \textit{ours w/o $\sigma$} and setting it to 0.1 as in \textit{MSc-iNCD}. The results in Tab. \ref{tab:MI} demonstrate both the improvement observed in \textit{MSc-iNCD+GA} and the performance decrease in \textit{ours w/o GA} underscore the positive impact of granular alignment on class discovery. Moreover, increasing the prototype count leads to improved performance, indicating that fine-grained modeling contributes to better clustering results. Besides, the decrease of moving $\sigma$ demonstrates the benefit of Gaussian density in modeling the probability rather than the common cosine distance.

\begin{wraptable}[13]{rT}{0.45\textwidth}
\begin{minipage}{0.45\textwidth}
\begin{table}[H] 
\scriptsize
\setlength{\tabcolsep}{5pt}
\vspace{-18pt}
\caption{Performance on knowledge preservation by different memory strategies.}
\vspace{-1.3mm}
\label{tab:Memory}
\setlength{\tabcolsep}{5pt}
\resizebox{\textwidth}{!}{
\begin{tabular}{lccccc}
\toprule
\multirow{2}{*}{Methods} & \multirow{2}{*}{\shortstack{Memory \\ Per Class}} & \multicolumn{2}{c}{C100} & \multicolumn{2}{c}{T200} \\
\cmidrule(lr){3-4} \cmidrule(lr){5-6}
& & $\mathcal{A}\uparrow$ & $\mathcal{F}\downarrow$ & $\mathcal{A}\uparrow$ & $\mathcal{F}\downarrow$ \\ \midrule
Examplar    & 5   & 56.1 & 39.2 & 44.7 & 37.4 \\
Examplar   & 20     & 69.8 & 24.4 & 60.3 & 19.5 \\
Examplar  & 50    & 76.1 & 15.9 & 65.4 & 14.3 \\
Class-mean    & 2 & 68.7 & 21.8 & 57.8 & 21.7 \\ \midrule
Ours (p500)   & $\approx$5/2.5      & 71.5 & 16.8 & 60.7 & 21.0 \\
Ours (p1000)  & $\approx$10/5        & 76.8 & 13.7 & 63.3 & 16.4 \\
Ours (p2000)  & $\approx$20/10        & 76.3 & 14.9 & 63.8 & 15.9 \\
Ours (p5000)  & $\approx$50/25        & \textbf{78.1} & \textbf{11.9} & \textbf{67.2} & \textbf{12.7} \\
\bottomrule
\end{tabular}
}
\end{table}
\end{minipage}
\end{wraptable}

\textbf{Fine-grained Prototypes for Knowledge Preservation.}
In addition, we highlight the advantages of the proposed fine-grained Gaussian prototype learning in retaining previous knowledge to prevent forgetting. As demonstrated in Tab. \ref{tab:Memory}, we present the overall discovery accuracy and forgetting score in comparison to 
other designed methods and only replace the memory strategies within our method.
\textit{Exemplar} represents a common strategy employed by CIL, where a random sample of exemplars \{5,20,50\} is selected for each discovered class and stored after each session.
\textit{Class-mean} is the strategy utilized in \textit{Msc-iNCD}, involving the computation and storage of the mean and variance for each discovered class.
For reference, we also report the performance using varying numbers of Gaussian prototypes from the set \{500, 1000, 2000, 5000\}, along with the corresponding memory size.
Tab. \ref{tab:Memory} shows that our strategy outperforms the practice of storing exemplars within the same memory. This advantage is due to our approach's capacity to capture and model the statistical nuances of feature distribution. Moreover, the result that employing more prototypes can lead to improved performance, further confirms the effectiveness of our method.

\begin{wraptable}[9]{rT}{0.45\textwidth}
\begin{minipage}{0.45\textwidth}
\begin{table}[H] 
\centering
\scriptsize
\setlength{\tabcolsep}{5pt}
\vspace{-12pt}
\caption{Overall accuracy and forgetting score on different overlap reduction strategies.}
\vspace{-1.0mm}
\label{tab:Joint}
\resizebox{\textwidth}{!}{
\begin{tabular}{lccccccc}
\toprule
\multirow{2}{*}{Methods} & \multicolumn{2}{c}{C100} & \multicolumn{2}{c}{T200} & \multicolumn{2}{c}{B200} \\ 
\cmidrule(lr){2-3} \cmidrule(lr){4-5} \cmidrule(lr){6-7}
 & $\mathcal{A}\uparrow$ & $\mathcal{F}\downarrow$ & $\mathcal{A}\uparrow$ & $\mathcal{F}\downarrow$ & $\mathcal{A}\uparrow$ & $\mathcal{F}\downarrow$  \\
 \midrule
KTRFR        & 75.3 & 14.1 & 62.1 & 17.5 & 45.9 & 5.3 \\
w/o Proj     & 70.0 & 12.5 & 63.1 & \textbf{11.5} & 45.5 & 2.3 \\
w/o $\mathcal{L}_{sep}$      & 73.7 & \textbf{12.1} & 62.1 & 15.1 & 46.2 & 2.4 \\
 \midrule
OURS & \textbf{76.8} & 13.7 & \textbf{63.3} & 16.4 & \textbf{50.1} & \textbf{2.1} \\
\bottomrule
\end{tabular}
}
\end{table}
\end{minipage}
\end{wraptable}

\textbf{Analysis of Overlap Reduction Strategy.}
As depicted in Tab. \ref{tab:Joint}, we examine the advantages of various training strategies incorporated into the overlap reduction method by selectively removing or replacing them. Specifically:
\textit{KTRFR} corresponds to the strategy employed in \textit{MSc-iNCD} \cite{liu2023MSciNCD}, involving classifier training through replaying old samples alongside new samples with pseudo labels. We report the performance when we replace our loss (Eq. \ref{eq:jointclassifier}) with \textit{KTRFR}
\textit{w/o $\mathcal{L}_{sep}$} represents the results after removing the separation loss in Eq. \ref{eq:seploss}.
\textit{w/o Proj} reflects the outcomes when removing the projector $g$.
The results emphasize the significance of each component, demonstrating their contributions to our final performance.

\vspace{-1mm}
\subsection{Hyper-Parameter Analysis}

\begin{wrapfigure}[20]{rT}{0.32\textwidth}
\begin{minipage}{0.32\textwidth}
\begin{figure}[H]
    \vspace{-22pt}
    \centering
    \subfigure{
        \includegraphics[width=0.92\textwidth]{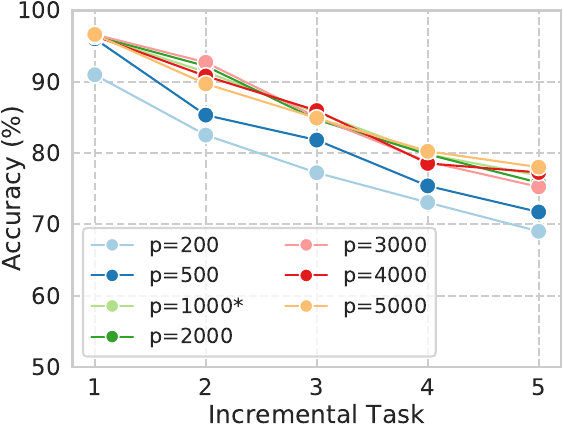}
    }
    \\[-0.5mm]
    \subfigure{
	\includegraphics[width=0.92\textwidth]{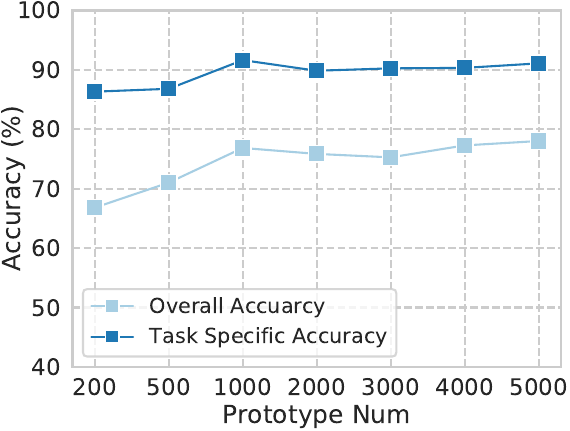}
    }
    \vspace{-2mm}
    \caption{Hyper-parameter analysis on the number of prototypes $pNum$.}
    \label{fig:pNum}
\end{figure}
\end{minipage}
\end{wrapfigure}

\textbf{Number of Fine-grained Prototypes.}
While we have briefly discussed the effects of adjusting the number of prototypes ($pNum$), this section provides a more detailed exploration of its impact. As shown in Fig. \ref{fig:pNum}, it becomes apparent that increasing the number of prototypes has a positive effect on both class discovery and forgetting reduction performance (b). Striking a balance between accuracy and memory size, we have selected $pNum=1000$ as the optimal choice for our method.

\begin{figure}[htp]
    \centering
    \subfigure{
        \includegraphics[width=0.24\textwidth]{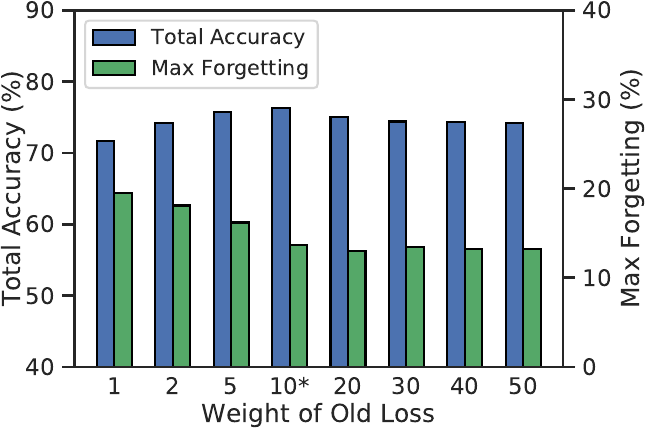}
    }
    \hspace{-3mm}
    \subfigure{
	\includegraphics[width=0.24\textwidth]{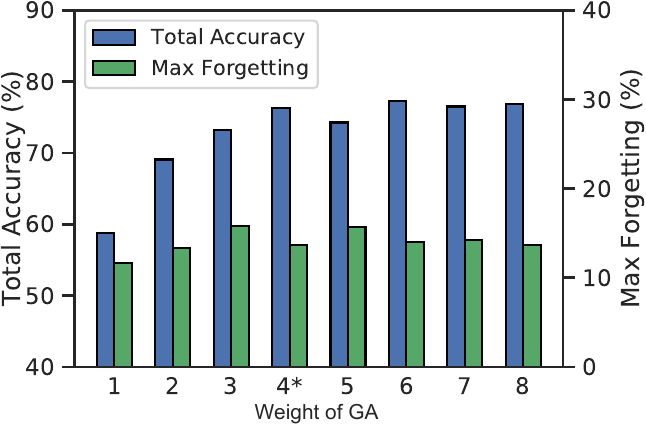}
    }
    \hspace{-3mm}
    \subfigure{
	\includegraphics[width=0.24\textwidth]{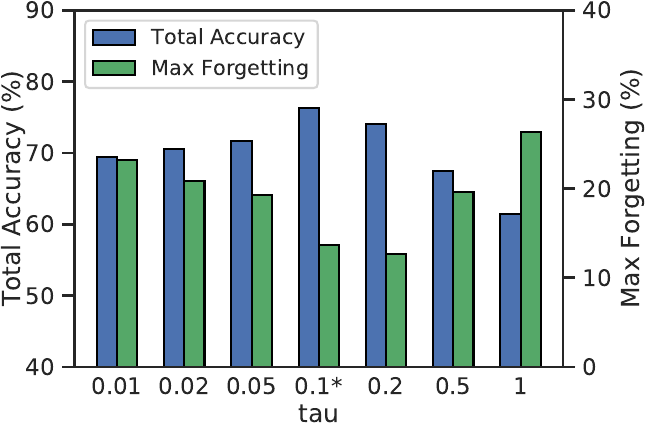}
    }
    \hspace{-3mm}
    \subfigure{
	\includegraphics[width=0.24\textwidth]{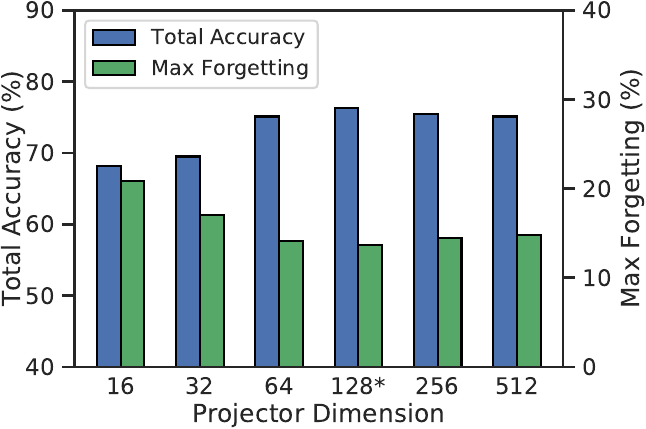}
    }
    \vspace{-2mm}
    \caption{Hyper-Parameter analysis on (a) the weight of old loss $\lambda_{old}$, (b) the weight of the granular alignment regularization term $\lambda_{GA}$, (c) temperature score $\tau$, and (d) output dimension of the projector.}
    \label{fig:param}
\end{figure}

\textbf{Other Parameters.}
As depicted in Fig. \ref{fig:param}, we analyze several other parameters employed in our method: (a) and (b) represent the performance when varying the weight of old loss $\lambda_{old}$ and the weight of granular alignment regularization term $\lambda_{GA}$, respectively. Notably, these weights exhibit stability across a wide range of values. (c) presents the sensitivity analysis of the temperature parameter $\tau$ utilized in Eq. \ref{eq: p(y|z)}. We opt for the recommended value of 0.1, consistent with most prior research employing $l_2$ normalization cosine distance. For the dimensions of the projector output,  (d) shows that our method performs consistently well across various commonly used settings. These results collectively underscore the robustness and non-sensitivity of our method to hyperparameter selection.

\vspace{-3mm}

\vspace{-4mm}

\section{Conclusion and Discussion}
\vspace{-1mm}
\textbf{Conclusion:} 
Our work delves into a more challenging and practical issue of unsupervised class incremental learning (UCIL). We analyzed that effectively capturing comprehensive feature representations and discovering unknown novel classes are critical to addressing this problem. Building on this insight, we propose a novel UCIL method by exploiting the fine-grained prototype distribution, along with the granularity alignment and overlap reduction strategy. The experiments have demonstrated significant improvements over existing state-of-the-art methods across five datasets. 

\textbf{Limitations \& Future work:} 
In this paper, we consider a more practical scenario compared to traditional CIL, where all labels are absent during training. However, practical situations may present more challenges, including the unknown number of classes, the intermixing of historical and novel classes, or imbalanced class distribution. Future research will strive to propose more effective methods to address these challenges in increasingly realistic scenarios.

\label{sec: limitations}

\bibliography{UCIL}
\bibliographystyle{splncs04}


\end{document}